%% file: main.tex
\documentclass[10pt,twocolumn,letterpaper]{article}

\usepackage{cvpr}
\usepackage{times}
\usepackage{epsfig}
\usepackage{graphicx}
\usepackage{amsmath}
\usepackage{amssymb}
\usepackage{verbatim}
\usepackage[pagebackref=true,breaklinks=true,letterpaper=true,colorlinks,bookmarks=false]{hyperref}

\usepackage{soul}
\usepackage[dvipsnames]{xcolor}
\usepackage{math_macro} 

\cvprfinalcopy 

\def\cvprPaperID{1620} 

\ifcvprfinal\fi
\begin{document}

\title{Novel Object Viewpoint Estimation through Reconstruction Alignment}

\author{Mohamed El Banani \quad Jason J. Corso \quad David F. Fouhey\\
University of Michigan\\
{\tt\small \{mbanani,jjcorso,fouhey\}@umich.edu}\\
}

\maketitle

\begin{abstract}
The goal of this paper is to estimate the viewpoint for a novel object. 
Standard viewpoint estimation approaches generally fail on this task due to their reliance on a 3D model for alignment or large amounts of class-specific training data and their corresponding canonical pose.
We overcome those limitations by learning a reconstruct and align approach.
Our key insight is that although we do not have an explicit 3D model or a predefined canonical pose, we can still learn to estimate the object's shape in the viewer's frame and then use an image to provide our reference model or canonical pose. In particular, we propose learning two networks: the first maps images to a 3D geometry-aware feature bottleneck and is trained via an image-to-image translation loss; the second learns whether two instances of features are aligned. 
At test time, our model finds the relative transformation that best aligns the bottleneck features of our test image to a reference image.
We evaluate our method on novel object viewpoint estimation by generalizing across different datasets, analyzing the impact of our different modules, and providing a qualitative analysis of the learned features to identify what representations are being learnt for alignment.
\end{abstract}


\section{Introduction}
\input{01_intro}

\input{F2_model_figure}

\section{Related Work}
\label{sec:related}
\input{02_related_work}

\section{Novel Object Viewpoint Estimation}
\label{sec:problem}
\input{03_problem}

\section{Approach}
\label{sec:approach}
\input{04_approach}

\section{Experiments}
\label{sec:experiments}
\input{05_results}

\input{06_discussion}

\vspace{6pt} \noindent
\textbf{Acknowledgments}
We would like to thank the reviewers and area chairs for their valuable comments and suggestions,
and the members of the UM AI Lab for many helpful discussions. 
Toyota Research Institute ("TRI") provided funds to assist the authors with their research but this article solely reflects the opinions and conclusions of its authors and not TRI or any other Toyota entity.

{\small
\bibliographystyle{ieee}
\bibliography{bibliography}
}

\end{document}

%% file: 01_intro.tex
Consider the two views of the owl in Figure 1. How are they related? As humans, we can easily imagine how to move from view 1 to view 2, even if we have never seen this object before. 
This problem of understanding object pose and viewpoint has long fascinated researchers in both computer vision (starting with the first PhD thesis on computer vision~\cite{Roberts65}) and psychology~\cite{shepard1971mental,Tarr1998}. 
Within computer vision, many approaches have been proposed to estimate an image's viewpoint with respect to a given, or assumed, oriented 3D model.
Despite the success of those approaches on known classes, they struggle with novel objects since both the 3D model and its canonical pose are unknown. 
In this work, we learn to predict the relative viewpoint for a novel object using a single reference view.

Historically, there has been two general approaches to viewpoint estimation.
First, 3D model alignment approaches find a transformation that aligns an image to a known 3D model~\cite{grimson1990object,pepik20123d,pascal3d}.
Despite their efficacy, those approaches were limited to objects with available 3D models.
Second, end-to-end discriminative approaches learn to directly estimate the image's viewpoint with respect to a canonical pose (\textit{e.g.}, head-on for a car)~\cite{su2015render, tremblay2018falling, tremblay2018dope, tulsiani2015viewpoints}.
While those approaches do not rely on explicit 3D models, their predictions are with respect to an implicitly-defined, canonically-oriented model. 
Due to their reliance on explicitly- or implicitly-defined 3D models, those approaches struggle with novel objects where neither the model nor its \textit{canonical} orientation are known.

\input{F1_teaser}

How can we estimate the viewpoint of a novel object? 
Viewpoint is defined with respect to a class-specific coordinate frame or a canonical pose (e.g., head-on for a car)~\cite{tulsiani2015viewpoints}.
For small generalization (\eg, from bikes to motorcycles), one can exploit similarity in viewpoint appearance between categories~\cite{kuznetsova2016exploiting}, but this is rarely the case. 
Motivated by research on mental rotation~\cite{shepard1971mental}, we observe that an image, as opposed to a 3D model, can serve as a reference pose.
Hence, instead of aligning to a 3D model, we find the relative transformation between two views of the same object.

In this work, we combine insights from both 3D alignment and end-to-end learning. 
We do not use a 3D model, but instead learn to map the input image to a 3D feature grid. 
Our key insight is that, although we do not have an explicit 3D model, 
we can still learn a deep network that maps each object instance to a 3D reference frame. 
This is done by combining learned 2D layers with projection layers constrained to follow projective geometry, similar to \cite{kar2017learning}. We then learn to align these 3D feature grids to identify the relative transformation that would minimize misalignment in the joint 3D feature grid.

We learn the 3D feature grid via 2D supervision (but not direct 3D supervision via voxels), while incorporating physical constraints inspired by 3D shape carving \cite{kutulakos2000theory}. 
We hypothesize that forcing the representation to pass through a geometry-aware bottleneck provides a useful inductive bias.
Further, the unlearned projection layers make it far easier for the network to learn an implicit 3D shape representation that can be inspected, where unstructured layers might lead to pure memorization that generalizes worse~\cite{tatarchenko2019single}. 
We then train an alignment network to estimate whether two views would produce an aligned 3D shape representation by training on randomly misaligned examples similar to~\cite{carreira2016human,li2016iterative}.

We evaluate novelty via generalization across different datasets. For instance, we show that our model can generalize well from ShapeNet~\cite{shapenet2015} to Thingi10K~\cite{Thingi10K} models, despite them looking very different as shown in Figure~\ref{fig:teaser} and the supplementary material. This sidesteps difficulties involved in using object classes as a proxy for novelty~\cite{kuznetsova2016exploiting, genre, xian2017zero}: 
class distinctions can be arbitrary, with some classes like motorcycles and bicycles sharing a lot of view-specific appearances.

We demonstrate the effectiveness of our proposed approach in Section \ref{sec:experiments}, and show that while standard approaches are challenged by novel objects, our approach generalizes gracefully to new datasets.
Specifically, our model achieves a viewpoint estimation accuracy of 40\% when generalizing from ShapeNet to Thingi10K, compared to 25\% accuracy achieved by standard approaches.
We run several analysis experiments to understand the learned representations, and find that our model also shows good across-dataset generalization on view prediction 
and voxel prediction; achieving an IoU score of 0.43 on ShapeNet airplanes despite being trained on Thingi10K. 
Overall, our experiments show that a reconstruction and alignment approach generalizes to novel objects better than discriminative approaches, and we hope that our work will spark more interest in this direction.

%% file: F1_teaser.tex
\begin{figure}[t]
\begin{center}
   \includegraphics[width=\linewidth]{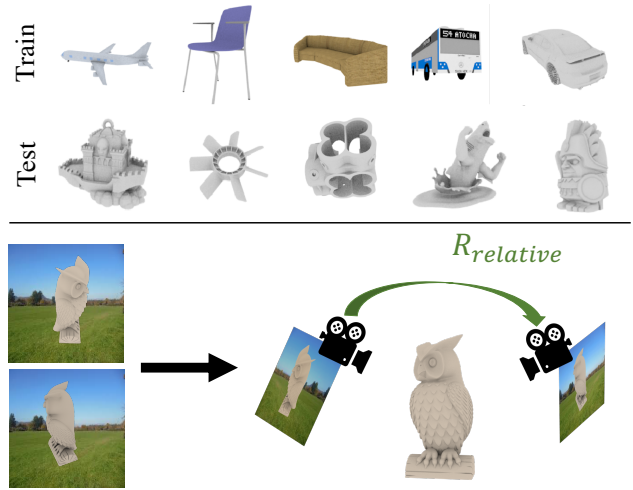}
\end{center}
   \caption{Humans can not help but see the 3D structure in those two views, which makes aligning their viewpoints very easy. This paper proposes a reconstruct-and-align approach to learning viewpoint estimation for novel objects.}
\label{fig:teaser}
\end{figure}

%% file: F2_model_figure.tex
\begin{figure*}[t]
    \begin{center}
    \includegraphics[width=\linewidth]{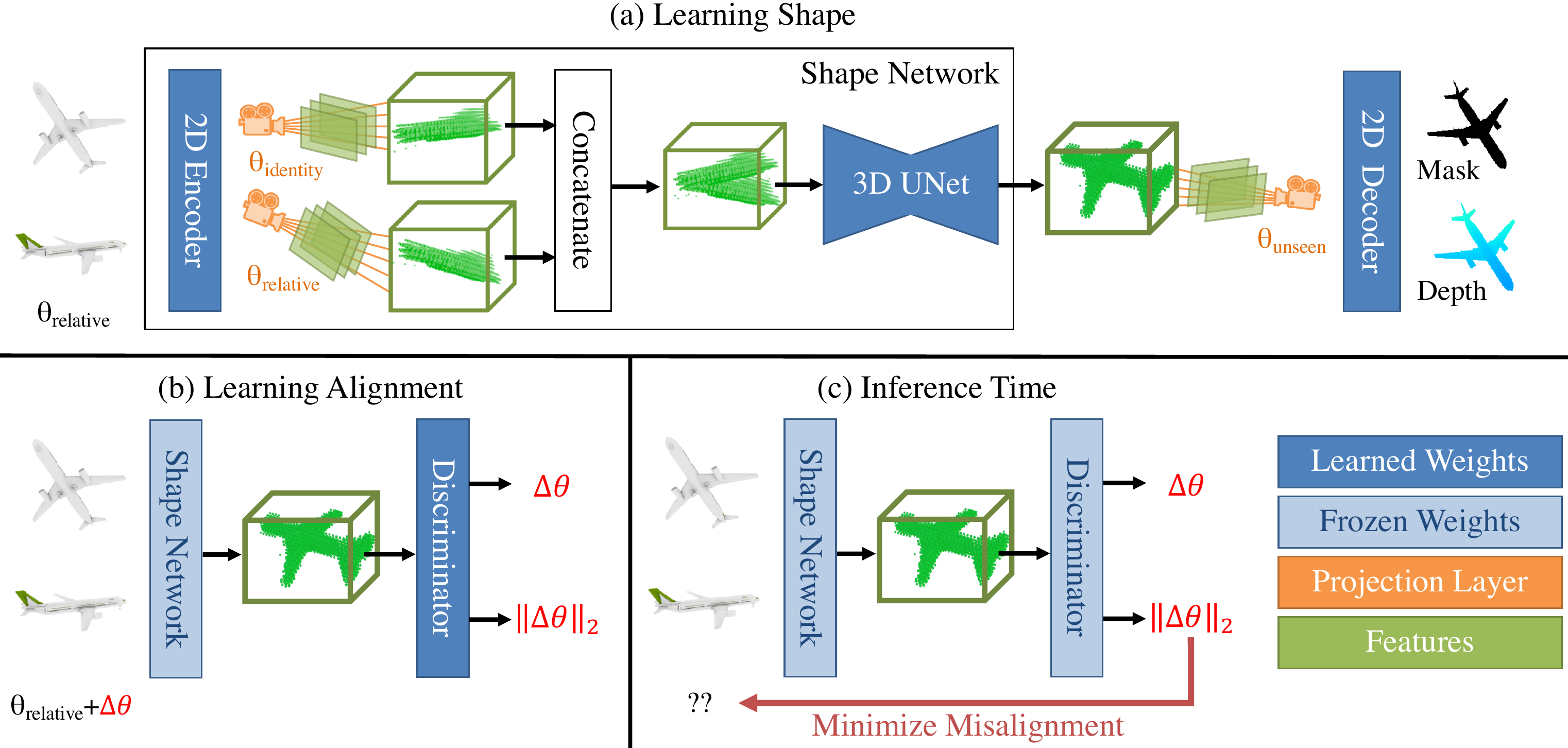}
    \end{center}
    \caption{\textbf{Approach Overview.} Our approach consists of two learning stages and an inference stage. 
    (a) We first learn to predict an object's shape from two views of the object and their relative pose. 
    We train our Shape Network to estimate the object's appearance from a third viewpoint. 
    (b) Using the trained Shape Network, we train a discriminator to predict the degree of misalignment between two views. 
    (c) At inference time, we find the relative pose that would best align the two inputs.
    Note: Input image backgrounds are removed for clarity.
    }
\label{fig:model}
\end{figure*}

%% file: 02_related_work.tex
The ability to accurately orient an object in 3D space in one of the oldest problems in computer vision~\cite{Roberts65}.
Early work posed the problem as an alignment problem, 
specifically that of estimating ``the transformation needed to map an object model from its inherent coordinate system into agreement with sensory data''~\cite{grimson1990object}. 
This formulation was powerful as it was task agnostic as long as one had access to the 3D model. 
Early approaches focused on finding correspondences between images and specific model instances~\cite{brooks1981symbolic, huttenlocher1990recognizing,lowe1987three}. 
More recent approaches focused on generalizing to all instances within an object classes by using class-specific 3D key-points~\cite{tseng2019fewshot,3dinterpreter, pascal3d, zhou2018starmap} or deformable part models~\cite{berg2005shape, pepik20123d}. 
While we are inspired by alignment approaches, we assume no access to a 3D model and generalize to novel object classes by learning to both reconstruct the 3D model and align images to it. 

Motivated by the success of large-scale image classification, a recent line of work focused on learning pose estimation with little or no supervision beyond image-pose pairs~\cite{ghodrati20142dinfo, mahendran20173d, su2015render, tremblay2018falling, tremblay2018dope, tulsiani2015viewpoints, xiang2018posecnn}. 
Those approaches replaced the alignment approach with fully-learned 2D feature-based models like convolutional neural networks (CNNs). 
Although this approach has been successful at synthetic-to-real generalization~\cite{su2015render, tremblay2018falling} and within-class generalization~\cite{ghodrati20142dinfo, tulsiani2015viewpoints}, 
it has remained unclear how successful it would be at generalizing to novel classes. 
While we also do not assume access to a 3D model; unlike this work, we train our model to be agnostic to object class, learn a 3D object representation, and generalize to novel classes. 

A recent line of work incorporates other losses or tasks with pose estimation such as 
object classification~\cite{kuznetsova2016exploiting, peng2015learning, pepik20123d, su2015render}, 
keypoint detection~\cite{tseng2019fewshot, tulsiani2015viewpoints, zhou2018starmap}, and 
object reconstruction~\cite{insafutdinov18pointclouds, tulsiani2018multiviewconsistency, yan2016perspective}. 
Closest to our approach are \cite{insafutdinov18pointclouds, tulsiani2018multiviewconsistency} who jointly learn 3D reconstruction and pose prediction from unannotated images. 
While our work stems from the same observation that pose and shape are closely related, our goals are different. 
While \cite{insafutdinov18pointclouds, tulsiani2018multiviewconsistency} are interested in learning shape and pose for a specific classes from very weak supervision, our focus is to generalize to novel classes of objects by learning from standard viewpoint supervision. 

Independent of this work, there has been a rising interest in learning more structured or meaningful intermediate representations through representational bottlenecks
~\cite{pepik20123d,tulsiani2015viewpoints,tung2018spatialcommonsense,wu2017marrnet,yang2015weakly,zhou2018starmap}. 
The representational bottleneck is interpreted as providing a good inductive bias to the learning process. 
More relevant to us are approaches that force a 3D representational bottleneck in the form of a voxel grid~\cite{kar2017learning, tulsiani2018multiviewconsistency, tung2018spatialcommonsense, yan2018learning, yan2016perspective}.
While some approaches use 3D supervision, others leverage the 3D representation bottleneck to learn 3D reconstruction using only 2D supervision.
We would like to emphasize that while we are inspired by this line of work, our goals are different; while those approaches focused on performing accurate single- or multi-view 3D reconstruction of a known class of objects, our goal is not to reconstruct the object, but rather to use the 3D bottleneck to allow us to estimate the relative viewpoint for novel objects.

Finally, our goal is to generalize to unseen classes. 
While this problem has been studied extensively in the scope of image classification (see ~\cite{xian2017zero} for a survey), it has not received a lot of attention for 3D tasks such as pose estimation. 
To the best of our knowledge, there has only been two previous approaches that tackled similar problems. 
Kuznetsova~\etal~\cite{kuznetsova2016exploiting} proposed a metric learning approach that performs joint classification and pose-estimation, and they leverage the learned space for zero-shot pose estimation. 
However, their approach only works if the novel object is similar to previously seen objects (\eg, cars to buses and bicycles to motorcycles), which limits the generalizability of their method. 
Tseng~\etal~\cite{tseng2019fewshot} proposed a keypoint alignment approach that learns to predict 3D keypoints to align the novel class. However, they expect 3D keypoint annotation for the reference images of the novel object at test time, while we only assume a single, unannotated, reference image. 


%% file: 03_problem.tex
Our overarching goal is to build a system that understands the {\it viewpoint} of a previously unseen object using as little information about the object as possible, and ideally generalizes between completely unrelated objects, such as zebras to forks instead of cars to buses. The difficulty is that viewpoint is defined with respect to a coordinate frame or canonical pose (\eg, head-on for a car): without it, the problem is fundamentally undefined and any viewpoint could be the origin viewpoint. 
Accordingly, past work has defined the coordinate frame using keypoints \cite{tseng2019fewshot}, or generalized across semantically similar objects \cite{kuznetsova2016exploiting} (\eg, cars to buses). 

Rather than use a predefined coordinate frame via keypoints or semantic similarity, we propose to instead use a single {\it image} as a reference. 
Our goal then is, given a reference image $\IB_1$ with viewpoint $\vB_1$ and an image $\IB_2$ whose viewpoint $\vB_2$ we would like to estimate, we want to predict the relative rotation
$\RB_{\textrm{relative}}$ such that $\vB_2 = \RB_{\textrm{relative}} \vB_1$.
This formulation maintains the essence of viewpoint estimation, while circumventing the limitations posed by requiring a canonical pose.
It should be noted given the relative pose between two images and the canonical pose of the object in one image, it is trivial to calculate the \textit{canonicalized} pose in the second image.

During training, we assume we have access to image-viewpoint pairs from an arbitrary number of classes. 
At test time, we are presented with two images from a new class and tasked with predicting the relative viewpoint of the second image with respect to the first image.

%% file: 04_approach.tex
The goal of this work is to build a system that can predict the relative viewpoint between two views of an object that it has never seen before. 
Our approach, shown in Figure~\ref{fig:model}, takes inspiration from the early formulation of pose estimation as alignment~\cite{grimson1990object}. 
However, instead of using a 3D model, we learn to predict a 3D representation from each view that can be aligned. 
We first present a high-level sketch of our approach before explaining each stage in more detail.
Architectural details are presented in Appendix A.1.
We analyze the effectiveness of our design choices in Section~\ref{sec:ablations}.

\vspace{6pt}
\noindent
\textbf{Approach Sketch.}
In the first stage, we \textit{learn shape} by training our model on a view prediction task. 
Given two images and the relative viewpoint between them, we train our model to extract 2D features from each image and back-project them into a joint 3D feature tensor.
During back-projection, we use the object's depth from each view to carve out the empty space between the viewer and the object.
For an arbitrary third view, we project the 3D feature tensor into 2D, and predict either the object mask or depth in that view. 
In the second stage, we \textit{learn alignment} by training a discriminator to predict if the extracted 3D feature tensor is well-aligned or not. 
At test time, we find the relative viewpoint by finding the transformation that minimizes predicted misalignment. 

\subsection{Learning Shape}

Given two images and a relative pose, $\{I_1, I_2, \RB_\text{relative}\}$, we train a Shape Network to generate a 3D feature grid, $\mathcal{F_\text{object}} \in \mathbb{R}^{F{\times}N{\times}N{\times}N}$, that captures the object's shape. 
Since we assume that the viewer is looking straight at the object for the first image, the second viewpoint is  $\RB_\text{relative}$.

The Shape Network consists of a 2D CNN encoder, a differentiable back-projection layer, and a 3D UNet~\cite{ronneberger2015unet,cciccek20163dunet}. 
We provide the specific architectural details in the supplementary material. 
The 2D encoder extracts a 2D feature map for each image that is then back-projected into a 3D voxel grid of dimension $N$ ($N$=32) using a viewpoint and a depth map. 
The back-projection is based on the differentiable projection layers from Kar \etal~\cite{kar2017learning}.
We back-project the features by projecting rays for each pixel location in the 2D feature map into the 3D grid using the appropriate viewpoint, the value for each voxel is then set via bi-linear sampling.

Inspired by early work on space carving~\cite{kutulakos2000theory}, we leverage known or estimated depth from the object to carve away the empty space in-front and around the object by setting the features in those voxels to 0.
After space carving, we have two feature grid tensors, $\mathcal{F}_{\text{v}_1}$,  $\mathcal{F}_{\text{v}_2}$, that capture the appearance and view of each object in the same frame of reference.

The two feature grid tensors from each view do not see the same parts of the objects and are, essentially, hallucinating the unseen parts. Therefore, we fuse them with the goal of predicting a consistent 3D representation of the object. We accomplish this by concatenating the two feature grids along the feature dimension to produce a tensor, $\mathbb{R}^{2F \times N \times N \times N}$, and pass it into a 3D UNet.
The 3D UNet's goal is to refine this representation by aligning the seen and hallucinating the unseen. 

We train the Shape Network on 2D view prediction of the object's mask or depth from a third view. 
This is done by projecting the 3D features to 2D using the relative viewpoint of the third view. 
Since we do not know the object's depth, we sample the features at multiple depth values along each ray. 
We use a $1\times1$ convolution to aggregate features for all sampled depth values for a given pixel location. 
This is followed by two convolutional layers and a single up-sampling layer in the middle to match the depth or mask output size. 
We train the network to minimize binary cross entropy from mask prediction, and $L_1$ loss for depth prediction.

\input{F3_mask_predictions}

\subsection{Learning Alignment} 

Once we have trained a Shape Network to estimate the object's 3D feature representation, our goal for the second stage is to use the 3D representation to predict whether two images of an object are aligned or not. 
We generate training data by freezing the Shape Network's weights and perturbing its input relative viewpoint by a random rotation. We then train a discriminator to predict the magnitude of that perturbation. 
Our intuition is that while determining the relative viewpoint between two objects might be difficult, misalignment is much easier to detect as it produces inconsistent object shapes, with the degree of inconsistency being a measure of how misaligned the images were.

We implement the discriminator as a 3D CNN. 
Our network architecture consists of two 3D inception modules followed by two fully-connected layers. 
We adapt the inception module~\cite{szegedy2015going} to 3D by expanding the kernels along the depth dimension.
We use the inception module as it allows the network to detect features at multiple scales which would be useful when detecting inconsistencies. 

We train the discriminator to predict the perturbation direction (Euler angles) and magnitude (geodesic distance). 
We randomly perturbed the relative viewpoint of two thirds of our instances by at least 10 degrees.
We train the network using an $L_2$ loss for all 4 outputs. 

\subsection{Relative Viewpoint Estimation} 
Given a trained Shape Network and a discriminator, the relative viewpoint is the viewpoint that will minimize predicted misalignment.  
This problem could be solved as an optimization problem or via a greedy search. 
In this work, we uniformly sample viewpoints and output the viewpoint that minimized the discriminator's predicted perturbation magnitude. 


%% file: F3_mask_predictions.tex
\begin{figure*}[t]
\begin{center}
   \includegraphics[width=\linewidth]{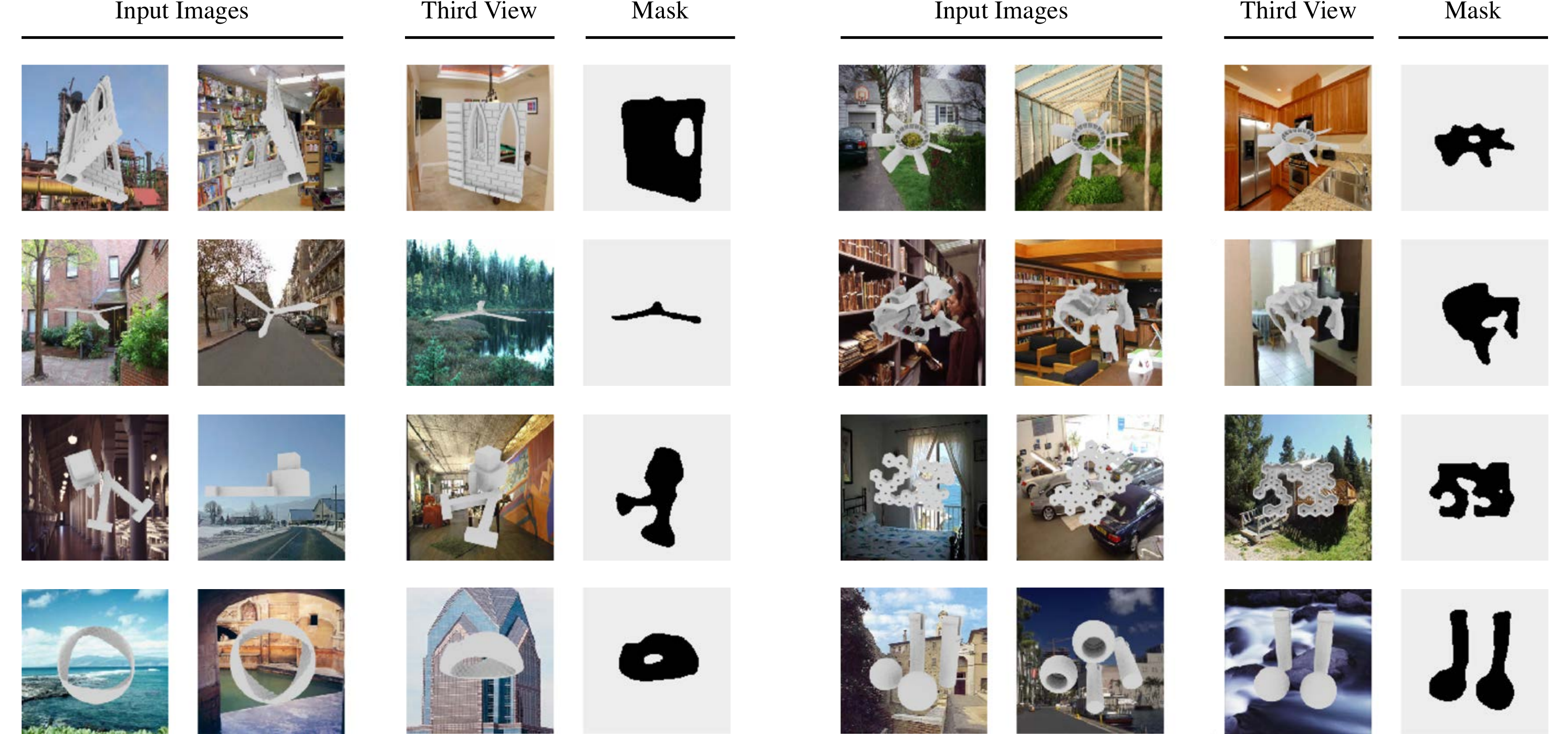}
\end{center}
   \caption{\textbf{Mask Prediction on Thingi10k.} Given two images, our models learns to predict the mask for a third view. The first two columns present two input views, while the third column is the view for which the mask is being predicted. The masks shown are predicted by a model which is trained on a different dataset; ShapeNet. We can accurately predict the 3D model's projection from a novel viewpoint despite being trained on a very different set of shapes.}
\label{fig:thingi_visuals}
\end{figure*}

%% file: 05_results.tex
We now empirically evaluate our model's performance on relative viewpoint estimation, as well as report additional experiments that analyze the different components of the model. 
Our experiments aim to answer two questions: (1) how well our proposed approach works, especially in the case of generalization to novel objects; and (2) how the method works and whether it has indeed learned to solve the task by using 3D representations. 

We address the first question by evaluating the model's ability to generalize across datasets for both viewpoint estimation (Section~\ref{sec:vp_results}) and view prediction (Section~\ref{sec:recon_results}). 
Cross-dataset generalization gives us a good proxy for how the model would perform on novel objects since ShapeNet and Thingi10K models look drastically different. 

We address the second question by evaluating our model's ability to learn shape through view prediction (Section~\ref{sec:recon_results}) and voxel prediction (Section~\ref{sec:voxels}). We also run several ablations to better understand the significance of different model components (Section~\ref{sec:ablations}).

\vspace{3pt}
\noindent
\textbf{Datasets.} 
We use three datasets: ShapeNet, Pix3D, and Thingi10K.
While ShapeNet and Pix3D are fairly uniform in terms of their models' types and appearances, Thingi10K models vary greatly in terms of their size, appearance, and geometric properties.
Thingi10K~\cite{Thingi10K} consists of $10K$ models created for 3D printing.
The lack of strong patterns or prototypical shapes makes Thingi10K very challenging for both pose estimation and reconstruction methods, and an excellent test dataset for our task.
Alternatively, ShapeNet's large size (55 categories and $\sim 57K$ models) makes it a great training set, while allowing us to evaluate our approach on models from the same class.
Finally, Pix3D~\cite{pix3d} is a smaller dataset, consisting of 395 3D models of 9 categories of furniture. 
The smaller number of models and domain-specificity of Pix3D provide an interesting comparison against the larger variance in object categories in ShapeNet and object shapes in Thingi10K.

\vspace{3pt}
\noindent
\textbf{Data Sampling.}
We choose our poses by uniformly sampling around a viewing sphere. With models centered around the origin, the pose transformation is simply the inverse of the viewpoint transformation. 
This deviates from previous work which sampled according to consumer photos' distributions to promote better synthetic to real generalization~\cite{su2015render}. 

After excluding models with missing or corrupted 3D model files, we end up with 55,281 ShapeNet models, 390 Pix3D models, and 9994 Thingi10K models.
For ShapeNet and Thingi10K, we sample 20 different poses per object, while we sample 200 views for Pix3D. We do so to mitigate the large difference between the numbers of models in those datasets. 
We randomly split our datasets by models to ensure that each of the splits has a disjoint set of models. 
We use 80\%/10\%/10\% of the data as our training, validation, and test splits.

\input{T1_viewpoint_table}

\input{05a_VP_Estimation}
\input{05b_2D_Recon}
\input{T3_ablations}
\input{05c_Ablations}
\input{05d_3d_results}


%% file: T1_viewpoint_table.tex
\begin{table*}[t]
\centering
\caption{\textbf{Relative Viewpoint Estimation Performance.}
We evaluate our model on its ability to perform within- and across-dataset generalization. 
The letters represents the initial letter of each dataset: $\mathcal{S}$hapeNet, $\mathcal{P}$ix3D, and $\mathcal{T}$hingi10k. 
Each column represents different train/test settings $\mathcal{S}\to\mathcal{T}$ is training on ShapeNet and testing on Thingi10k.
We find that our model is able to generalize across datasets, while previous approaches demonstrate a clear bias towards their training distributions.}
\label{tab:exp_pose_synthetic}
\begin{tabular}{|c | c c  |  c c |  c c | c c  | c c | c c | }
\hline
   
                            & \multicolumn{2}{c|}{$\mathcal{S} \to \mathcal{S}$} 
                            & \multicolumn{2}{c|}{$\mathcal{P} \to \mathcal{P}$} 
                            & \multicolumn{2}{c|}{$\mathcal{T} \to \mathcal{T}$} 
                            & \multicolumn{2}{c|}{$\mathcal{S} \to \mathcal{P}$} 
                            & \multicolumn{2}{c|}{$\mathcal{S} \to \mathcal{T}$} 
                            & \multicolumn{2}{c|}{$\mathcal{P} \to \mathcal{T}$} 
                            \\ 
\hline 
                            & $Acc_{\pi/6}$     & MedErr      
                            & $Acc_{\pi/6}$     & MedErr      
                            & $Acc_{\pi/6}$     & MedErr      
                            & $Acc_{\pi/6}$     & MedErr      
                            & $Acc_{\pi/6}$     & MedErr     
                            & $Acc_{\pi/6}$     & MedErr      
                            \\

\hline 
Stat. Prior             & 4.15  & 114.71    & 4.13  & 115.01    & 4.33  & 114.41    & 4.13 & 115.01     &  4.33 & 114.41    & 4.33  & 114.41    \\
RPNet                   & \textbf{60.23} & \textbf{44.32}     & \textbf{68.93} & \textbf{32.61}     & 20.82 & 91.43     & \textbf{66.42 }& 46.07     & 25.00 & 91.85     & 14.43 & 111.80    \\
ConvAE                  & 50.49 & 52.65     & 51.96 & 56.36     & 15.35 & 100.56    & 48.63 & 46.78     &  8.65 & 116.11    &  3.55 & 129.29    \\
Ours                    & 47.57 & 66.04     & 58.08 & 46.02     & \textbf{41.25} & \textbf{61.96}     & {60.67} & \textbf{43.47}     & \textbf{40.12} & \textbf{70.08}     & \textbf{33.00 }& \textbf{96.82}     \\
\hline 
\end{tabular}
\end{table*}

%% file: 05a_VP_Estimation.tex
\subsection{Viewpoint Estimation}
\label{sec:vp_results}

We first evaluate how well our system performs on viewpoint estimation, our primary task. Our aim here is to evaluate how well the system performs, especially when generalizing from one dataset to another.

\vspace{6pt}
\noindent\textbf{Experimental Setup. }
We evaluate pose estimation performance based on the geodesic distance between the predicted and ground-truth pose as was done in previous work~\cite{su2015render, tulsiani2015viewpoints}. Given two rotation matrices, the geodesic distance provides the magnitude of the transformation between those two rotation matrices. Following ~\cite{huynh2009metrics}, we calculate geodesic distance as $D(R_{gt}, R_{pr}) = ||\log{\mathbf{R}_{gt}\mathbf{R}^\top_{pr}}||$. Following~\cite{su2015render, tulsiani2015viewpoints}, we report the median geodesic distance and the percentage of viewpoints under threshold $\pi/6$. 

\vspace{6pt}
\noindent\textbf{Baselines.}
We compare our models performance to several baselines. 
For all our baselines, we use binned Euler angles to represent rotation and train the models using a cross-entropy loss. 

\vspace{2pt}
\noindent\textbf{\textit{Statistical Prior.}} We find the mode pose for the validation and use it on testing. Since we uniformly sampled the viewpoints, we find that this baseline is very weak and corresponds to a random baselines. 

\vspace{2pt}
\noindent\textbf{\textit{RPNet.}} En \etal~\cite{en2018rpnet} proposed a Siamese-network to perform relative camera pose estimation. 
As reported by the authors, we find that for more difficult datasets, it is better to train each network separately on absolute viewpoint and then calculate the relative viewpoint explicitly. 
We used binned Euler angles instead of quaternions as we found that they perform better.
Furthermore, we provide their network with RGB-D input to provide a fair comparison to our method's use of depth-based space carving.

\vspace{2pt}
\noindent\textbf{\textit{Convolutional Autoencoder (ConvAE)}.} 
Given that our model learns both reconstruction and pose estimation, we trained a convolutional autoencoder on both tasks.
We initially used UNets~\cite{ronneberger2015unet} due to their success in image translation tasks, but found that removing the skip connections improved the viewpoint estimation performance.
Viewpoint is estimated on the encoded features using a 3-layer CNN. 
Like RPNet, we found that training on absolute viewpoint improved performance.

\paragraph{Results. }

We find that directly estimating viewpoint scales poorly do datasets like Thingi10K. As shown in Figure~\ref{fig:thingi_visuals}, Thingi10K models exhibit very high variability which makes them challenging for a model that maps 2D patterns directly to a pose as seen by in Table~\ref{tab:exp_pose_synthetic}. 
By contrast, our model is able to leverage this high variability in the data as it allows it to learn better reconstruction features. This is evident by the large performance gain that our model on Thingi10K; both when training and testing on it, or when transferring to it from another dataset.

On the other hand, we find that if there is limited difference between training and testing (e.g., training and testing on ShapeNet), directly estimating viewpoint does well. Interestingly, we find that in this domain, training on more data with higher variance improves performance as seen with training on ShapeNet and evaluating on Pix3D, especially since all of Pix3D's classes exist in ShapeNet. This pattern is shown both for the direct pose models as well as our approach. 
Another interesting finding is that joint reconstruction and pose estimation do not explain our models performance as shown by the convolutional autoencoder baseline. It appears that the additional reconstruction task burdens the network as opposed to providing it with an additional learning signal. 

We note that RPNet is representative of current end-to-end discriminative viewpoint-estimation models as they generally follow the structure of a CNN backbone and FC layers to predict viewpoint~\cite{mahendran20173d, su2015render, tremblay2018dope, tulsiani2015viewpoints, xiang2018posecnn}.
As a result, the comparison to RPNet represents a more general comparison against problems that frame viewpoint estimation as a discriminative task. 

Furthermore, recent approaches proposed jointly learning reconstruction and pose estimation, where the network learns to predict pose in an \textit{emergent} coordinate frame~\cite{insafutdinov18pointclouds, tulsiani2018multiviewconsistency}. 
While those models are trained to minimize a reconstruction error, they are still using a CNN to directly estimate the pose with respect to a class-specific, \textit{emergent} canonical pose. Hence, they face the same challenges as RPNet. Additionally, those approaches can perform better when given ground-truth pose, which we provide to RPNet.

%% file: 05b_2D_Recon.tex
\input{T2_reconstruction_table}

\subsection{2D Reconstruction}
\label{sec:recon_results}

We now analyze our Shape Network's ability to represent shape, as well as how well that ability generalizes to novel objects. 
Specifically, we evaluate the Shape Network's ability to predict the object's mask or depth from an unseen viewpoint.
Note that our model is trained on either depth or mask prediction. Hence, the depth and mask results are shown for models trained on the respective task.
Similar to the previous experiment, we are interested in the model's ability to generalize to unseen classes. 
We emphasize that we have a disjoint set of training and testing models, hence, the testing performance on the same dataset evaluates class-specific generalization, while testing across datasets evaluates generalization to unseen classes. 

\vspace{8pt} \noindent
\textbf{Experimental Setup. }
We train and test the performance on the 9 pairwise dataset pairs for ShapeNet, Pix3D, and Thingi10K. 
We evaluate mask reconstructions using IoU at a threshold of 0.5 as well as a pixel-level F1 score. 
Following~\cite{eigen2014depth}, we evaluate depth estimation using log root mean-square-error and threshold accuracy, $\delta_{1.25}$. 
For a depth prediction, $y$, threshold accuracy is calculated as the percentage of pixels, $y_i$, s.t. $\max(\frac{y_i}{y^{gt}_i}, \frac{y^{gt}_i}{y_i}) < 1.25$.

\vspace{8pt} \noindent
\textbf{Results. }
We find that our model achieves high performance for mask prediction across all the datasets as shown in Table~\ref{tab:exp_2drecon_synthetic}. 
More importantly, we find that there is very little variance between the model's performance across datasets. This demonstrates our model's ability to learn to represent 3D shape in a generalizable way . 
This is in stark contrast to recent findings from reconstruction that find that 3D reconstruction approaches will often memorize some of the training models~\cite{tatarchenko2019single}. 
We hypothesize that our model is less susceptible to memorization due to (a) the geometry-aware bottleneck and (b) the lack of a shared canonical pose across class instances. 
The geometry-aware bottle neck forces the model to learn to represent its input in 3D. 
Furthermore, by removing the canonical pose assumption, the model can no longer rely on specific locations in the tensor having a consistent representation (\eg, airplane wings always being on the side). 
As a result, the model is forced to represent the specific input, as opposed to just classifying it.

Of course, like any learned approach, our model is still affected by the quality of the training data. We observe that, in general, the best performing model on a dataset is the one trained on that dataset with two interesting exceptions. 
The first exception is that for mask prediction, ShapeNet does very well on Pix3D. This is likely due to the degree of class overlap between ShapeNet and Pix3D, as well as the additional variety of models seen in ShapeNet.
The second exception is that Pix3D trained model tend to do better on depth prediction. While it is not clear why this would happen, one potential explanation is that the planarity of most furniture might provide a good scaffolding for the networks to learn better depth, while other datasets provide too much variance for the network to learn good representations. 

%% file: T2_reconstruction_table.tex
\begin{table*}[!t]
\centering
\caption{\textbf{2D Reconstruction Cross-dataset Generalization Experiment}. The rows show the training dataset, while the columns present the test dataset. We find that our model can easily generalize across datasets, with very minor performance drops. This supports our claim that our model is learning a generalizable representation of shape, as opposed to memorizing the shapes available within each dataset.
\vspace{1pt}} 
\label{tab:exp_2drecon_synthetic}
\begin{tabular}{|c | c c | c c |  c c | c c | c c | c c | }
\hline
   
                    & \multicolumn{4}{c|}{ShapeNet} 
                    & \multicolumn{4}{c|}{Pix3D} 
                    & \multicolumn{4}{c|}{Thingi10k} \\
\hline
                    & \multicolumn{2}{c|}{\textit{Mask}} & \multicolumn{2}{c|}{\textit{Depth}} 
                    & \multicolumn{2}{c|}{\textit{Mask}} & \multicolumn{2}{c|}{\textit{Depth}} 
                    & \multicolumn{2}{c|}{\textit{Mask}} & \multicolumn{2}{c|}{\textit{Depth}} 
                    \\ 
                    & IoU   & F1    & LogRMSE & $\delta_{1.25}$     & IoU   & F1    & LogRMSE & $\delta_{1.25}$    & IoU   & F1    & LogRMSE & $\delta_{1.25}$
                    \\

\hline 
ShapeNet     & \textbf{0.83}  & \textbf{0.90}    & \textbf{0.53}  & 0.66             & \textbf{0.80 } &\textbf{ 0.88}    & 0.63  & 0.60                           & 0.82  & 0.90          & 0.35  & 0.84\\
Pix3D        & 0.78  & 0.87                      & 0\textbf{.53}  & \textbf{0.74}    & 0.79  & \textbf{0.88}             & \textbf{0.59} & \textbf{0.74  }       & 0.79  & 0.87          & 0.34  & \textbf{0.90}\\
Thingi10k    & 0.76  & 0.85                      & 0.69  & 0.70                      & 0.73  & 0.84                      & 0.81 & 0.63                           & \textbf{0.85}  & \textbf{0.91}          & 0\textbf{.30  }& 0.88 \\
\hline 
\end{tabular}
\end{table*}

%% file: T3_ablations.tex
\begin{table}[t]
\centering
\caption{\textbf{Ablation Study.} We ablate our model by removing two key components in the 3D and analyzing the effect on 2D reconstruction and relative viewpoint estimation.}
\label{tab:exp_ablations}
\begin{tabular}{|l | c c |c | c c | c |}
\hline
   
                            & \multicolumn{3}{c|}{$\mathcal{S} \to \mathcal{S}$} 
                            & \multicolumn{3}{c|}{$\mathcal{S} \to \mathcal{T}$} 
                            \\ 
\hline
                     & $Acc_{\pi/6}$     & MedErr    & IoU       & $Acc_{\pi/6}$     & MedErr    & IoU 
                            \\
\hline 
Ours           &  47.57            & 66.04     & 0.83      & 40.12             & 70.08     & 0.82  \\
\hline 
- carving            & 6.38              & 134.23    & 0.64      &  4.21             & 130.14    & 0.63  \\
- refine             & 47.57             &  59.76    & 0.63      & 42.16             &  54.03    & 0.64  \\
\hline 
\end{tabular}
\end{table}

%% file: 05c_Ablations.tex
\subsection{Ablations}
\label{sec:ablations}

We perform an ablation study to better understand the model's performance and failure modes. 
In particular, we ablate two central elements of our model: 3D refinement and space carving.
We train our ablated models on ShapeNet and evaluate them on both ShapeNet and Thingi10K. 
We follow the same training procedure as before. The results are shown in Table~\ref{tab:exp_ablations}. 

\vspace{6pt} \noindent
\textbf{Space Carving.}
We find that removing space carving is extremely detrimental to the model. 
The model's 2D reconstruction performance suffers, while the viewpoint performance is reducing to random guessing.
Space carving appears to greatly boost our performance by providing a class and model agnostic benefit. 
Given two projected views, space carving provides a high degree of certainty to the model regarding which voxels can be ignored, and hence, greatly improves the alignment. 
The uncertainty about the object's extent appears to stifle its ability to align views. 

\vspace{8pt} \noindent
\textbf{3D Refinement.}
We observe that removing the 3D refinement network also greatly deteriorates the reconstruction performance. 
However, unlike space carving, removing it actually boosts the viewpoint performance. 
This is likely caused by the network hallucination of features (in order to complete the object), which in turn is mixed with the non-hallucinated features, which may hamper alignment.

%% file: 05d_3d_results.tex
\subsection{Features to Voxels}
\label{sec:voxels}

Given that the model learns to project to and from a voxel grid, we expect that the features of this grid can capture occupancy despite never having been trained on 3D information. 
While we do not ever directly extract an occupancy grid during training, we find that we can adapt one of the layers in the network to generate it. 
Specifically, while training the Shape Network, we add a multi-view consistency loss~\cite{tulsiani2018multiviewconsistency, drcTulsiani17}. 
In particular, Tulsiani~\etal~\cite{tulsiani2018multiviewconsistency} propose a loss that samples different points along projected rays to detect ray termination events.
We extend their formulation to a feature grid by defining a linear layer that maps the ray-sampled features to a permeability score.
Since the permeability layer maps each feature vector to a ray termination probability, it correlates with the occupancy likelihood. 
Hence, we apply it to the feature tensor output from the Shape Network to generate an occupancy grid.
We note that removing this loss does not significantly change our viewpoint estimation or 2D reconstruction performance.

\begin{figure}[t]
\begin{center}
   \includegraphics[width=\linewidth]{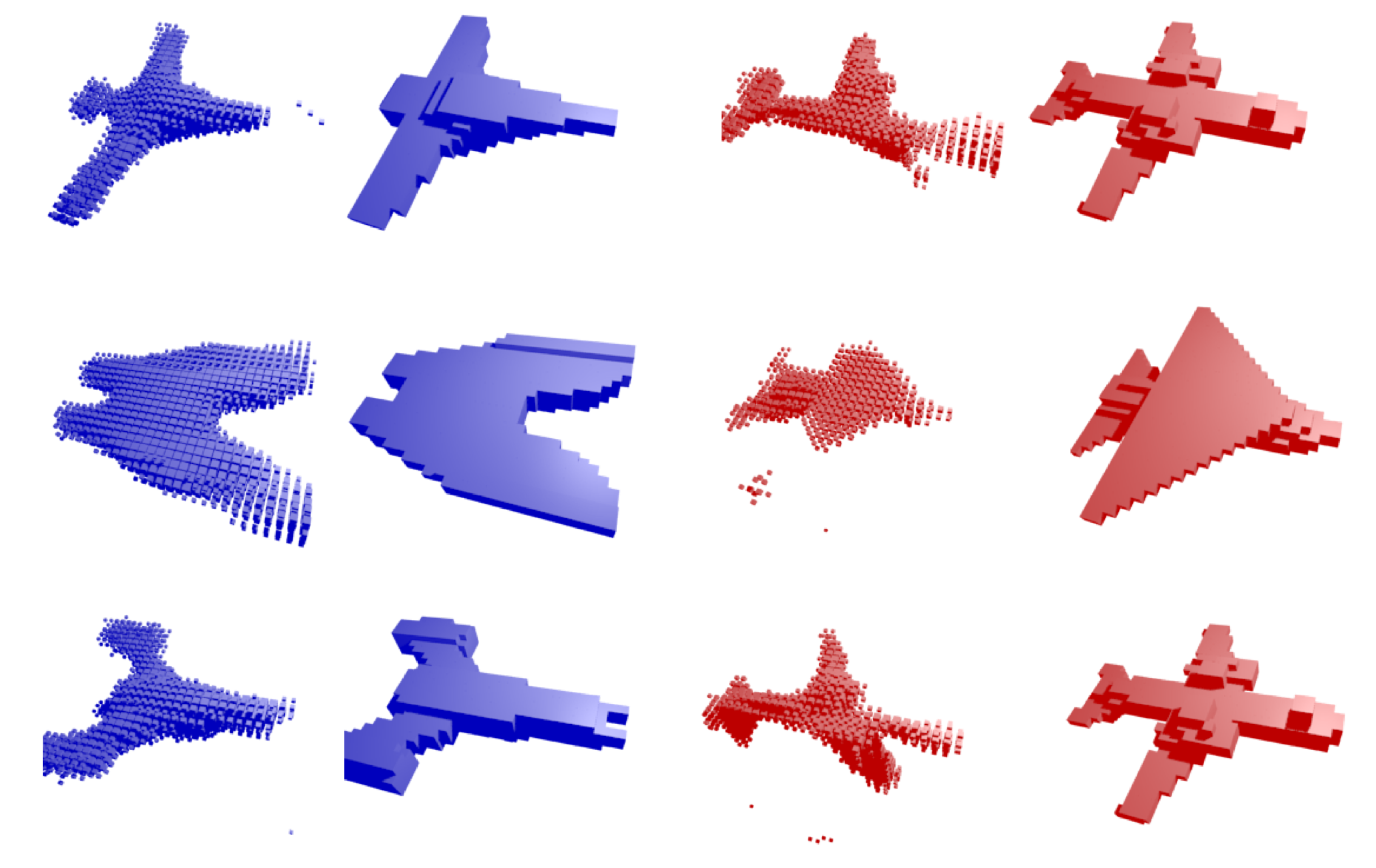}
\end{center}
   \caption{\textbf{Features To Voxels.} Our model can accurately predict occupancy without being trained using voxels or for voxel prediction. We present 3 success cases (IoU $>$ 0.7;~ in blue) and 3 failure cases of our model (IoU $<$ 0.1; in red), with our prediction on the left and the ground-truth on the right. We observe that our failure cases tend to arise from failure to complete an occluded part or registering the part at an incorrect location, which would be expected from a model that has never seen a plane before.}
\label{fig:voxels}
\end{figure}

We train our model on Thingi10K and evaluate it on the airplane class of ShapeNet. We would like to emphasize that Thingi10K does not include an airplane class, so we are evaluating on unseen objects. 
We use the voxel grids provided by ShapeNet, but we center and resize them to fit a voxel grid of dimension 32. 
We evaluate the quality of our rendering by calculating the intersection over union (IoU) of the occupancy grids. We pick the threshold based on the validation set.

Since our embedding is in a viewer-centric frame of reference, we rotate it back to the canonical frame before computing IoU. Our model achieves a mean IoU of 0.43 with 98\% of the samples having an IoU at least 0.25.
We note that this is calculated on the direct embedding dimension without performing any registration or scaling to better match ShapeNet 3D models. We visualize several success and failure cases in Figure~\ref{fig:voxels}.

%% file: 06_discussion.tex
\section{Conclusion}

In this paper, we present a reconstruction and alignment approach for novel object viewpoint prediction. 
We observe that previous viewpoint prediction approaches have either relied on large amounts of class-specific, canonically-oriented data or on having a 3D model for alignment. 
However, those two assumptions do not apply when dealing with a novel object. 
Our key insight is that while we may not have access to a 3D model, we can learn a deep network that estimates the object's shape in the viewer's frame of reference, and use a reference view to determine the viewpoint with respect to other views of the object. 
To this end, we propose learning two networks: the first network uses a 3D feature bottleneck to represent objects in 3D; the second network learns whether or not two views are aligned. 

We evaluate our approach on several datasets. We find that our approach significantly outperforms standard viewpoint estimation approaches when there is a large domain shift between the train and test models, or when the training models themselves have high variance.  Furthermore, we find that we can extract 3D occupancy grids despite not training using 3D supervision. This provides ample evidence that our models can generalize to unseen objects.

Our current models greatly rely on the depth input to accurately reconstruct the object. 
While the 2.5D input might be needed to understand the 3D structure of a novel object, we would like to explore ways of minimizing this reliance in future work. Furthermore, we would like to explore optimization approaches that would allows us to converge to the best viewpoint instead searching over a set of predetermined viewpoints.